# Reasoning at the Right Time Granularity

**Suchi Saria**        **Uri Nodelman**        **Daphne Koller**
Department of Computer Science
Stanford University
Stanford, CA 94305
{ssaria,nodelman,koller}@cs.stanford.edu

## Abstract

Most real-world dynamic systems are composed of different components that often evolve at very different rates. In traditional temporal graphical models, such as dynamic Bayesian networks, time is modeled at a fixed granularity, generally selected based on the rate at which the fastest component evolves. Inference must then be performed at this fastest granularity, potentially at significant computational cost. Continuous Time Bayesian Networks (CTBNs) avoid time-slicing in the representation by modeling the system as evolving continuously over time. The expectation-propagation (EP) inference algorithm of Nodelman et al. (2005) can then vary the inference granularity over time, but the granularity is uniform across all parts of the system, and must be selected in advance. In this paper, we provide a new EP algorithm that utilizes a general cluster graph architecture where clusters contain distributions that can overlap in both space (set of variables) and time. This architecture allows different parts of the system to be modeled at very different time granularities, according to their current rate of evolution. We also provide an information-theoretic criterion for dynamically re-partitioning the clusters during inference to tune the level of approximation to the current rate of evolution. This avoids the need to hand-select the appropriate granularity, and allows the granularity to adapt as information is transmitted across the network. We present experiments demonstrating that this approach can result in significant computational savings.

## 1 Introduction

Reasoning about systems that evolve over time is an important task that arises in many applications. The standard representational frameworks for temporal reasoning — hidden Markov models (Rabiner & Juang, 1986) and dynamic Bayesian networks (Dean & Kanazawa, 1989) — model the system by slicing time into a sequence of equal-length intervals. However, many systems are comprised of components that change on vastly different time scales. When modeling a person's activity in an office environment, some factors, such as their current job specification, evolve quite slowly, others, such as their current project composition, at a medium rate, and yet others, such as their current immediate task, evolve very quickly (often depending on what email they happened to get in the past few minutes). Similar high disparities in time granularity occur when modeling complex geopolitical situations, a person's television viewing pattern, and many more.

The framework of *continuous time Bayesian networks (CTBNs)* (Nodelman et al., 2002) provides a representation for structured dynamic systems that avoids the use of a fixed time granularity. CTBNs build on the framework of homogeneous Markov processes (Norris, 1997), which provide a model of continuous-time evolution. CTBNs model each process variable as a continuous-time Markov process, whose dynamics depends on other process variables in the model. Thus, not only can variables evolve at different rates, but the evolution rate of a single process variable can vary over time, in response to events occurring elsewhere in the system.

Exact inference in CTBNs involves generating an exponentially-large matrix representing the transition model over the entire system state. Nodelman et al. (2005) present an approximate inference algorithm for CTBNs which is an instance of the *expectation propagation (EP)* algorithm (Minka, 2001). In this algorithm, the system is segmented into time intervals that can vary in their length; within each segment, messages are passed in an EP cluster graph, which contains clusters that represent distributions over subsets of variables during that segment. While the time segments can be of different length, all the clusters of variables are broken up over the same segment boundaries. Thus, if one cluster evolves more rapidly than others, requiring a finer-grained approximation, inference in the entire system will have to be approximated at that granularity. Moreover, the granularity needs to be selected by the user, in advance, a design choice which is far from obvious.

In this paper, we present a new EP-based algorithm that has two important novel features. First, the algorithm uses a flexible cluster graph architecture where clusters, and messages between them, can have varying time scopes. This feature allows us to fully exploit the natural time-granularity at which different sub-processes evolve by modeling different parts of the system at different time granularities. Second, we introduce a new *dynamic-EP* algorithm, where the algorithm dynamically chooses the appropriate level of granularity to use in each cluster at each point in time. This level can depend both on the current evidence for that subset and on messages received from other parts of the system. Dynamic-EP utilizes an information-theoretic criterion to automatically decide whether a cluster should be encoded at a finer time granularity, and also how it should be split. We illustrate the performance of our Dynamic-EP algorithm on networks where different variables evolve at different rates; our results suggest that Dynamic-EP provides a much better time-accuracy trade-off than using a uniform granularity.

## 2 Continuous Time Bayesian Networks

We begin by briefly reviewing the key definitions of Markov processes and continuous time Bayesian networks, as presented by Nodelman et al. (2002).

A finite state, continuous time, homogeneous Markov process $X$ with state space $Val(X) = \{x_1, \ldots, x_n\}$ is essentially a distribution over a continuum of *transient variables* $X^t$ for $t \in [0, \infty)$. It is described by an initial distribution $P_X^0$ and an $n \times n$ *transition intensity matrix* whose off-diagonal entries $q_{x_i x_j}$ encode the intensity of transitioning from state $x_i$ to state $x_j$ and whose diagonal entries $q_{x_i} = \sum_{j \neq i} q_{x_i x_j}$. This matrix describes the transient behavior of $X^t$. If $X_0 = x$ then it stays in state $x$ for an amount of time exponentially distributed with parameter $q_x$. Upon transitioning, $X$ shifts to state $x'$ with probability $q_{xx'}/q_x$. If $P_X^0$ is the distribution over $X$ at time 0, then the distribution over the state of the process $X$ at some future time $t$ can be computed at $P_X^t = P_X^0 \exp(\mathbf{Q}_X \cdot t)$, where exp is matrix exponentiation.

A *continuous time Bayesian network* (CTBN) $\mathcal{N}$ defines a distribution over trajectories $\sigma$ for a set of process variables $\mathbf{X}$. A complete trajectory $\sigma$ can be represented as a sequence of states $\mathbf{x}_i$ of $\mathbf{X}$, each with an associated duration. A CTBN encodes this distribution in a factored form, as follows: Each process variable $X$ is associated with a *conditional Markov process* — an inhomogeneous Markov process whose intensity matrix varies as a function of the current values of a set of discrete conditioning variables $\mathbf{U}$. It is parameterized using a *conditional intensity matrix* (CIM) — $\mathbf{Q}_{X|\mathbf{U}}$ — a set of homogeneous intensity matrices $\mathbf{Q}_{X|\mathbf{u}}$, one for each instantiation of values $\mathbf{u}$ to $\mathbf{U}$. A *continuous time Bayesian network* $\mathcal{N}$ over $\mathbf{X}$ consists of two components: an *initial distribution* $P_{\mathbf{X}}^0$, specified as a Bayesian network $\mathcal{B}^0$ over $\mathbf{X}$, and a *continuous transition model*, specified using a directed (possibly cyclic) graph $\mathcal{G}$ whose nodes are $X \in \mathbf{X}$; $\mathbf{U}_X$ denotes the parents of $X$ in $\mathcal{G}$. Each variable $X \in \mathbf{X}$ is associated with a conditional intensity matrix, $\mathbf{Q}_{X|\mathbf{U}_X}$. The CIMs can be combined to form a single homogeneous Markov process over the joint state space by a process of *amalgamation*.

The resulting density over complete trajectories can be formulated within the framework of exponential families (see Lauritzen (1996)). We define a sufficient statistics vector $\tau(\sigma)$ comprised of sufficient statistics for each variable $\{T[x|\mathbf{u}], M[x, x'|\mathbf{u}]\}$: $T[x|\mathbf{u}]$ — the amount of time that $X = x$ while $\mathbf{U}_X = \mathbf{u}$; and $M[x, x'|\mathbf{u}]$ — the number of times that $X$ transitions from $x$ to $x'$ while $\mathbf{U}_X = \mathbf{u}$. Similarly, the natural parameters for the $\mathbf{Q}_X$ component of the model are simply the diagonal terms and the logarithm of the off-diagonal terms $\eta(\mathbf{Q}_{X|\mathbf{u}}) = \{-q_{x|\mathbf{u}}, \ln(q_{xx'|\mathbf{u}})\}$. Then, the probability of the trajectory over the variables in the model can be written as the inner product of the sufficient statistics and the natural parameter vectors:

$$P_{\mathcal{N}}(\sigma) \propto \prod_{X \in \mathbf{X}} P_X^0 L_X(T[X|\mathbf{U}], M[X|\mathbf{U}])$$

$$L_X(T[X|\mathbf{U}], M[X|\mathbf{U}]) = \prod_{\mathbf{u}} \exp(\langle \tau_{X|\mathbf{u}}(\sigma), \eta(\mathbf{Q}_{X|\mathbf{u}})\rangle) =$$

$$\exp\left(\sum_{\mathbf{u}} \sum_x -q_{x|\mathbf{u}} T[x|\mathbf{u}] + \sum_{\mathbf{u}} \sum_{x' \neq x} M[x, x'|\mathbf{u}] \ln(q_{xx'|\mathbf{u}})\right)$$

The term $L_X(T[X|\mathbf{U}], M[X|\mathbf{U}])$ is $X$'s *likelihood contribution* to the overall probability of the trajectory.

## 3 Expectation Propagation for CTBNs

We want to compute answers to probabilistic queries given some partial observations about the current trajectory. Although many forms of evidence are possible, we focus, for simplicity of presentation, on *interval evidence* of the form "Process variable $X$ takes the value $x$ for the duration of an interval $[t_1, t_2]$". From here on, we represent the composite of such evidence for all variables in the system in the interval $[t_1, t_2]$ as $\sigma^{t_1:t_2}$.

In this section, we briefly review the algorithm of Nodelman et al. (2005) (NKS from now on), which forms the basis for our approach. The NKS algorithm is based on the expectation propagation framework, which performs message passing in a cluster graph. In general, a cluster graph is defined in terms of a set of clusters $\mathcal{C}_j$, whose *scope* is some subset of the variables $\mathbf{V}_j \subseteq \mathbf{X}$. Messages are passed between clusters along edges $\mathcal{C}_j$—$\mathcal{C}_k$, each of which is associated with a *sepset* $\mathcal{S}_{j,k}$ whose scope is the set of variables $\mathbf{V}_{jk} = \mathbf{V}_j \cap \mathbf{V}_k$. The NKS algorithm uses a cluster graph whose clusters correspond to subsets of process variables over a particular time interval $[t_1, t_2]$; the cluster $\mathcal{C}_j$ encodes a distribution over the *trajectories* of the variables $\mathbf{V}_j$ during $[t_1, t_2]$, i.e., a distribution over the continuum of transient variables $X^t$, where $X \in \boldsymbol{V}_j$ and $t \in [t_1, t_2]$. A sepset $\mathcal{S}_{j,k}$ is used to transmit a distribution over the trajectories of the variables $\boldsymbol{V}_{jk}$ in the intersection of the two clusters. To pass a message from $\mathcal{C}_j$ to $\mathcal{C}_k$, the distribution in $\mathcal{C}_j$ is marginalized over the variables in the sepset, and the resulting marginal is passed to $\mathcal{C}_k$.

**Example 3.1** *Consider a chain CTBN $A \to B \to C \to D$, and an initial distribution $P^0_{ABCD} = P^0_A P^0_B P^0_C P^0_D$. The natural cluster graph for this CTBN has the structure $AB$—$BC$—$CD$. The $AB$ cluster, for example, could be initialized to contain the CIMs $\mathbf{Q}_A, \mathbf{Q}_{B|A}$ and $P^0_A P^0_B$. To pass a message from $AB$ to $BC$ over the sepset $B$, we would compute the distribution over $AB$ trajectories in the cluster, marginalize to produce a distribution over $B$, and pass the message to the $BC$ cluster. Importantly, although the joint $AB$ distribution is a homogeneous Markov process over $AB$, the marginal distribution over $B$ is not typically a homogeneous Markov process.*

In general, the exact marginal distribution over a subset of the variables in the network can be arbitrarily complex, requiring a number of parameters which grows exponentially with the size of the network. NKS address this problem by using *expectation propagation* (EP) (Minka, 2001). EP is a general scheme for approximate message passing in a cluster graph, where one approximates a complex message $\delta_{j \to k}$ by projecting it into some fixed parametric form in the exponential family, keeping the complexity of the messages bounded. The projection is selected to minimize the KL-divergence between $\delta_{j \to k}$ and its approximation $\hat{\delta}_{j \to k}$. In the NKS algorithm, the exponential family used for the message representation is the class of homogeneous Markov processes, characterized by an initial distribution and an intensity matrix.

So far, we have described a cluster graph where all clusters are over a fixed time interval $[t_1, t_2]$. To address the general case, NKS string together a sequence of cluster graphs, over consecutive intervals $[t_1, t_2], [t_2, t_3], \ldots$. Messages are passed from one interval $[t_i, t_{i+1}]$ to another $[t_{i+1}, t_{i+2}]$ by computing a point distribution at the boundary point $t_{i+1}$. In this solution, the intervals $[t_i, t_{i+1}]$ can have different lengths, but all of the clusters within a single cluster graph for an interval have precisely that interval as their time scope. This assumption can be costly in cases where some clusters evolve much more rapidly than others. In this case, the messages associated with a rapidly evolving cluster cannot be approximated well using a single homogeneous Markov process. To obtain a high-accuracy result, we must refine the representation of this cluster to utilize a much finer time granularity. However, this would force us to refine all clusters in the graph in a similar way, including clusters that evolve much more slowly. This refinement can greatly, and unnecessarily, increase the number of messages required. Moreover, the NKS algorithm requires the discretization of the clusters to be determined in advance, a design decision which is far from trivial, and provide no automated way to select or refine the clusters.

## 4 Variable Time-Scope EP

In this paper, we propose an alternative cluster graph architecture, which allows us to much more flexibly tune our approximation to the rate of evolution of each cluster separately. In our framework, each cluster $\mathcal{C}_j$ encodes a distribution over a set of process variables $\boldsymbol{V}_j$ over a *cluster-specific* interval $\mathcal{I}_j = [t^j_1, t^j_2]$. A cluster can encode the distribution over its interval using one or more piece-wise homogenous Markov processes. The sepset between $\mathcal{C}_j$ and $\mathcal{C}_k$ is a distribution over (a subset of) the intersection between the transient variables in the two clusters: the variables in $\boldsymbol{V}_{jk}$ over the interval $\mathcal{I}_j \cap \mathcal{I}_k$. The sepset always stores its distribution as a single homogenous continuous-time Markov process. Note that for two consecutive clusters $\mathcal{C}_j, \mathcal{C}_k$ over the same set of variables $\boldsymbol{Y}$, where $t^j_2 = t^k_1$, the sepset is simply the point distribution over $\boldsymbol{Y}$ at time $t^j_2$.

In this generalized cluster graph, even within the same time period, some clusters can span much longer intervals, whereas others can be much shorter. This flexibility allows the outgoing information regarding one set of process variables to be approximated fairly coarsely, using a message sent via a single sepset, whereas the information about others can be approximated in a much finer-grained representation by sending a message over multiple sepsets over the interval.

**Example 4.1** *Consider the CTBN of Example 3.1, over a time interval $[0, 6]$. We might choose a cluster graph that has: $\mathcal{C}_1$ with scope $A, B$ over the interval $[0, 6]$; $\mathcal{C}_2$ and $\mathcal{C}_3$ with scope $B, C$ and intervals $[0, 2]$ and $[2, 6]$; and $\mathcal{C}_4, \mathcal{C}_5, \mathcal{C}_6$ with scope $C, D$ and intervals $[0, 1]$, $[1, 3]$, and $[3, 6]$. Between each of these uniform clusters, we have a sepset $\mathcal{S}_{2,3}$ over the single time point 2, and two sepsets $\mathcal{S}_{4,5}$ and $\mathcal{S}_{5,6}$ over the time points 1 and 3 respectively. In addition, we have six sepsets that connect different variable scopes: $\mathcal{S}_{1,2}$ and $\mathcal{S}_{1,3}$ with scopes $B$ and intervals $[1, 2]$ and $[2, 6]$ respectively; $\mathcal{S}_{2,4}, \mathcal{S}_{2,5}, \mathcal{S}_{3,5}$ and $\mathcal{S}_{3,6}$ with scopes $C$ and intervals $[0, 1]$, $[1, 2]$, $[2, 3]$ and $[3, 6]$ respectively. Thus, the information that the $B, C$ clusters receive about the $A, B$ clusters is summarized within a single homogeneous Markov process; the information about the $C, D$ clusters is actually a piece-wise homogeneous Markov process with three separate segments, potentially providing a more precise approximation.*

Based on this general scheme, we now describe the details of the algorithm and the message passing steps that it takes. For the duration of this discussion, consider a particular CTBN $\mathcal{N}$ with an initial distribution specified as a Bayesian network $\mathcal{B}^0$ and a set of CIMs $\mathbf{Q}_{X|\mathbf{U}_X}$. We are interested in a particular time interval $[0, T]$, and may have some partial evidence $\sigma^{0:T}$ about the trajectory, as described above.

**Cluster Graph Construction.** Our message passing algorithm applies to a very general form of a cluster graph. As described above, each cluster $\mathcal{C}_j$ has a scope of variables $\boldsymbol{V}_j$ and an associated time interval $[t_1^j, t_2^j]$. In addition, clusters are related by sepsets, which also have a variable scope and a time scope. Importantly, the same pair of clusters can be related by more than one sepset. We use $\mathcal{S}_{j,k}^l$ to enumerate the sepsets relating the clusters $\mathcal{C}_j$ and $\mathcal{C}_k$; we use $\boldsymbol{V}_{jk}^l$ to denote the variable scope of $\mathcal{S}_{j,k}^l$ and $\mathcal{I}_{jk}^l$ its interval time scope. A legal cluster graph must satisfy the following three properties. **Family preservation**: For each CIM $\mathbf{Q}_{X|\mathbf{U}_X}$ and each time point $t$ there must exist some $\mathcal{C}_j$ such that $\boldsymbol{V}_j \supseteq (\{X\} \cup \mathbf{U}_X)$ and $[t_1^j, t_2^j] \ni t$; this allows the CIM to be placed in $\mathcal{C}_j$ at time $t$. Similarly, for each conditional probability distribution $P(X^0 \mid \mathbf{U}^0)$ in the time 0 Bayesian network $\mathcal{B}^0$, there must exist $\mathcal{C}_j$ such that $\boldsymbol{V}_j \supseteq (\{X\} \cup \mathbf{U})$ and $t_1^j = 0$. **Sepset containment**: For each sepset $\mathcal{S}_{j,k}^l$ relating a pair of clusters $\mathcal{C}_j, \mathcal{C}_k$, we have that $\boldsymbol{V}_{jk}^l \subseteq \boldsymbol{V}_j \cap \boldsymbol{V}_k$ and $\mathcal{I}_{jk}^l \subseteq \mathcal{I}_j \cap \mathcal{I}_k$. Moreover, $\cup_l \mathcal{I}_{jk}^l = \mathcal{I}_j \cap \mathcal{I}_k$. **Running intersection**: For each transient variable $X^t$, the set of clusters and sepsets containing $X^t$ forms a tree.

In most cases, the clusters will have a uniform structure over variable scopes across time, as in Example 4.1; we have a (non-disjoint) partition of the process variables into subsets, and a sequence of consecutive clusters for each such subset. However, this formulation also allows a different breakdown of process variables into clusters at different points in time, which is useful in cases where the strength of the interaction between variables can vary over time.

We now initialize the cluster graph. To understand this step, we first need to examine closely the forms of the measures that we obtain in a cluster $\mathcal{C}_j$ by aggregating the factors assigned to the cluster, the incoming messages, and the evidence. Recall that our evidence gives us observations of the form $X = x$ over an interval $[t_1, t_2]$. During that interval, we must *reduce* the system dynamics to consider only states consistent with $X = x$ (see NKS for details). Thus, if our cluster does not have uniform observation throughout the length of its interval, different time segments will have different dynamics. Different dynamics can also arise from messages. Recall that each sepset $\mathcal{S}_{j,k}^l$ sends the cluster $\mathcal{C}_j$ a message over an interval $\mathcal{I}_{jk}^l = [t_1, t_2]$, which is a sub-interval of $\mathcal{I}_j$. Thus, the distribution over $\mathcal{C}_j$ is broken up into a sequence of closed sub-clusters, representing consecutive sub-intervals of $\mathcal{I}_j$, each of which has a single coherent model for the system dynamics over its interval. Between each pair of consecutive sub-intervals $(t_{ji}, t_{j(i+1)})$ and $(t_{j(i+1)}, t_{j(i+2)})$, we have a *demarcation point* $t_{j(i+1)}$. Thus, the demarcation points are: the beginning or end points of any interval of evidence; the beginning or end points of any interval of an adjoining sepset $\mathcal{S}_{j,k}^l$; and the beginning and end of the entire interval $\mathcal{I}_j$.

We can now define a cluster's distribution in terms of these demarcation points and the closed intervals between them. To motivate the parameterization for each closed interval, we recall from NKS the recursive definition for computing probability distributions over variable states in a homogenous Markov process. For example, we compute a point distribution for $\boldsymbol{X}$ at $t$ given partial evidence $\sigma^{0:T}$ as: $P(\boldsymbol{X}^t|\sigma^{0:T}) = \frac{1}{Z}P(\boldsymbol{X}^t|\sigma^{0:t_1})\exp(\mathbf{Q}(t-t_1))\boldsymbol{\Delta}_{\boldsymbol{x},\boldsymbol{x}}\exp(\mathbf{Q}(t_2-t))P(\sigma^{t_2:T}|\boldsymbol{X}^t)$. Here, $Z$ is the normalizing constant representing the probability of the evidence, $\boldsymbol{\Delta}_{\boldsymbol{x},\boldsymbol{x}}$ is a zero matrix with a 1 in the row and column that both correspond to $x$; $t_1$ and $t_2$ are time points within $[0, T]$. Thus, for each closed interval $[t_1, t_2]$ within a cluster $\mathcal{C}_j$, to allow for efficient recursive computations, we maintain a data-structure that caches these components:

1. $P(\boldsymbol{V}_j^t|\sigma^{0:t_1})$ accessed as $\pi_j[\alpha^{t_1}]$
2. CIM $\mathbf{Q}_{\boldsymbol{V}_j}$ accessed as $\pi_j[\mathbf{Q}^{t_1}]$
3. $P(\sigma^{t_2:T}|\boldsymbol{V}_j^t)$ accessed as $\pi_j[\beta^{t_2}]$.

Each sepset contains the same data-structure as above represented using the same notation. We denote the message in $\mathcal{S}_{j,k}^l$ as $\mu_{j,k}^l$. A demarcation point contains messages exchanged between consecutive closed sub-intervals within a cluster, similarly represented as $\mu_j^t$ for the point at time $t$ in $\mathcal{C}_j$. Note, both demarcation points and sepsets between clusters of the same variable scope do not contain CIMs (i.e., they only contain distributions over their point intervals).

To initialize the cluster graph data-structures, we begin by setting all point distributions, $\alpha$ and $\beta$ vectors, to 1. We also initialize all CIMs to zero over their scope. Now, each factor from $\mathcal{B}^0$ is multiplied to $\alpha^0$ in a cluster thats starts at time 0 and contains the factor's variable scope. Moreover, the CIMs are assigned to a cluster such that, for each $t$ and each $X$, $\mathbf{Q}_{X|\mathbf{U}_X}$ is present in the cluster graph exactly once. In the case of uniformly structured cluster graphs, we simply pick, for each $X$, a single cluster sequence whose scope contains $X$ and $\mathbf{U}_X$, and incorporate $\mathbf{Q}_{X|\mathbf{U}_X}$ into each cluster in the sequence. After all the CIMs have been assigned, each interval may contain zero or more CIMs assigned to it. We store any evidence about the variables in the scope of a cluster interval, by reducing the CIMs and the corresponding point distributions. The resulting reduced matrices, which we call *dynamics matrices*, have the same form as intensity matrices, but do not necessarily satisfy the constraint that the diagonal entries are the negative sum of the off-diagonal entries. The dynamics matrices for each interval are amalgamated to produce a single dynamics matrix for that interval that describes the evolution of the variables in that interval. The amalgation and reduction operations are carefully detailed in NKS. To briefly recap, amalgamation corresponds to addition of the intensity matrices (after they have been expanded to apply to the same variable scope). Reduction corresponds to zeroing out the elements in the intensity matrix that are inconsistent with the evidence.

**Example 4.2** *Continuing Example 4.1, assume that we ob-*

serve $B = b$ during $[4, 5]$. The distribution at $C_1$ will have the following demarcation points: the beginning (1) and end (6) of the interval; the beginning (4) and end (5) of the interval of evidence over $B$; and the point 2 which is both the end of the sepset $S_{1,2}$ and the beginning of the sepset $S_{1,3}$. To obtain the dynamics matrix for the interval $(4, 5)$, for example, we amalgamate the CIMs $\mathbf{Q}_{B|A}$ and $\mathbf{Q}_A$ assigned to it, and reduce them to match the evidence $B = b$. Also, $\pi_1[\alpha^4]$ and $\pi_1[\beta^5]$ are reduced to match the evidence.

**Message Passing.** Given an initialized cluster graph, we now iteratively pass messages between clusters, until convergence. Convergence occurs when messages between neighboring clusters cease to affect the potentials. At a high level, during message passing each cluster $C_j$ collects incoming messages from all of the sepsets. Each such message is a distribution over the scope of the sepset. The messages are combined together with the factors stored at the cluster, and conditioned on the evidence. The result is an implicit description of a complex, unnormalized measure over the trajectories of the variables $\mathbf{V}_j$ throughout the interval $\mathcal{I}_j$. We now perform inference within the cluster to compute outgoing messages over each of the cluster's sepsets. This message is the cluster distribution, marginalized over the sepset scope, and projected into the parametric form of the outgoing message.

**Message Computation.** Our task now is to compute an outgoing message over a sepset $S_{j,k}^l$, with variable scope $\mathbf{V}_{jk}^l$ and interval $\mathcal{I}_{jk}^l$. When the variable scopes of $C_j$ and $C_k$ are not the same, we call a message exchanged between them a *vertical message*. To compute the outgoing message, we need to perform inference over the measure at $C_j$, and then perform a KL-divergence minimizing projection into the space of homogenous Markov processes over $\mathbf{V}_{jk}^l$ and $\mathcal{I}_{jk}^l = [t_1, t_2]$. Consider first the simple case where the entire cluster has uniform dynamics $\mathbf{Q}_j$, and so the cluster boundaries exactly match the sepset boundaries, and the only demarcation points are at the beginning and end of the interval, where we have factors $\phi_{\mathbf{V}_j}^{t_1}$ and $\phi_{\mathbf{V}_j}^{t_2}$, respectively. The message computation and marginalization operation is now identical to that described by NKS.

Recapping briefly, for any pair of instantiations $\mathbf{x}, \mathbf{x}'$ to $\mathbf{V}_j$, define $\boldsymbol{\Delta}_{\mathbf{x},\mathbf{x}'}$ to be a matrix of the same size as $\mathbf{Q}_j$ with zeros everywhere except for a 1 in the row corresponding to $\mathbf{x}$ and the column corresponding to $\mathbf{x}'$. As described by NKS, we can compute the projection of the distribution into the space of homogeneous Markov processes by computing the expected sufficient statistics:

$$\mathbf{E}[T[\mathbf{v}]] = \frac{1}{Z} \int_{t_1}^{t_2} \phi^{t_1} \exp(\mathbf{Q}_j(t-t_1)) \boldsymbol{\Delta}_{\mathbf{x},\mathbf{x}} \exp(\mathbf{Q}_j(t_2-t)) \phi^{t_2} \, dt$$

$$\mathbf{E}[M[\mathbf{x},\mathbf{x}']] = \frac{q_{\mathbf{x}\mathbf{x}'}}{Z} \int_{t_1}^{t_2} \phi^{t_1} \exp(\mathbf{Q}_j(t-t_1)) \boldsymbol{\Delta}_{\mathbf{x},\mathbf{x}'} \exp(\mathbf{Q}_j(t_2-t)) \phi^{t_2} \, dt$$

The normalization constant $Z$ in both equations is the partition function, which makes the expected amount of time over all states sum to $t_2 - t_1$. We can calculate all of these statistics simultaneously for all $\mathbf{x}, \mathbf{x}'$, using a fifth order Runge-Kutta numerical integration method with an adaptive step size.

In the general case, a cluster may have multiple sub-intervals. Note that the boundaries of a sepset always correspond to demarcation points in its neighboring clusters. If the sepset interval $[t_1, t_2]$ is a single sub-interval in $\mathcal{I}_j$, then the sufficient statistics computation follows exactly as described above. If $[t_1, t_2]$ spans multiple intervals, then for each interval $[t_{ji}, t_{j(i+1)}] \in [t_1, t_2]$, we repeat the above computation using the factors in $\pi_j^{[t_1:t_2]}$ which are the set of factors in $C_j$ corresponding to the interval $[t_1, t_2]$. Having computed all the sub-interval sufficient statistics, we now sum up these sufficient statistics over the different sub-intervals to obtain the overall set of sufficient statistics for the sepset interval. We marginalize the sufficient statistics over the smaller subset $\mathbf{V}_{jk}^l$, and compute the KL-divergence minimizing projection for the sepset intensity matrix by matching moments as follows: $q_{\mathbf{y}\mathbf{y}'} = \frac{\mathbf{E}[M[\mathbf{y},\mathbf{y}']]}{\mathbf{E}[T[\mathbf{y}]]}$, $q_{\mathbf{y}} = \sum_{\mathbf{y}' \neq \mathbf{y}} q_{\mathbf{y}\mathbf{y}'}$. Summing up the sufficient statistics over the sub-intervals has natural meaningful semantics. If the sufficient statistics over the different sub-intervals are the same, then a computation of the intensity matrix for each of the sub-intervals would result in the same intensity matrix. Hence, using a piece-wise homogenous Markov process parameterization would contain the same information as a single homogenous Markov process over the entire interval. Alternatively, if the sufficient statistcs vary widely between the different sub-intervals, then projection into a single homogenous Markov process would lead to a poor approximation.

Finally, to compute the point distributions, $\mu_{i,j}^l[\alpha^{t_1}]$ and $\mu_{i,j}^l[\beta^{t_2}]$ at $t_1$ and $t_2$ for the message, we simply marginalize the distributions at those time points in the cluster.

**Message Incorporation.** We will now discuss how to incorporate a message from the sepset. Again, consider the simple case where the incoming message boundary overlaps a single sub-interval within a cluster. We use factor multiplication and division to incorporate each component of the message into $C_k$.

When the incoming message spans multiple sub-intervals within the receiving cluster, we first represent the message in the form of the receiving cluster as a set of messages over its sub-intervals. For example, say $[t_{k0}, t_{k1}], \cdots, [t_{ki}, t_{k(i+1)}], \cdots, [t_{k(n-1)}, t_{kn}] \subseteq [t_1, t_2]$ are the corresponding sub-intervals in $C_k$. For each sub-interval $[t_{ki}, t_{k(i+1)}]$, we compute $\delta_{j \to k}^l[\alpha^{t_{ki}}]$, $\delta_{j \to k}^l[\beta^{t_{k(i+1)}}]$ and $\delta_{j \to k}^l[\mathbf{Q}^{t_{ki}}]$ from the newly-received message $\delta_{j \to k}$. We also compute $\mu_{j,k}^l[\alpha^{t_{ki}}]$, $\mu_{j,k}^l[\beta^{t_{k(i+1)}}]$ and $\mu_{j,k}^l[\mathbf{Q}^{t_{ki}}]$ from the old-message $\mu_{j,k}^l$ in the sepset. We obtain all of the above point distributions using standard forward and backward propagation within the interval. The sub-interval **Q** matrices are obtained by replicating the bigger interval matrix over each sub-interval. Now, the message for each sub-interval is incorporated independently.

To maintain consistency between adjacent clusters that are over the same variable scope and have an overlapping time point, we send what we call a *horizontal message* between them. Finally, as new information is received through messages that are incorporated in sub-intervals within a cluster, this information must be propagated along the remaining cluster intervals to maintain consistency between sub-intervals. This is similar to maintaining consistency across cluster boundaries since, essentially, we can view the sub-intervals as defining a chain-structured cluster graph, embedded in the larger cluster $C_j$ (as in the nested junction tree of Kjaerulff (1997)). We can now pass messages over this embedded cluster graph using an exact message passing algorithm.

**Summary**. We formally outline below the three forms of message passing steps we just discussed:

**Procedure** *Send-Vertical-Message*$(j, k, S_{jk}^l)$

1. $[t_1\ t_2] \leftarrow \mathcal{I}_{jk}^l$
2. $\delta_{j \to k} \leftarrow \text{marg}_{C_j \setminus V_{jk}^l}^{[t_1\ t_2]}(\pi_j)$
3. foreach sub-interval $[t_{ki}, t_{k(i+1)}] \subseteq [t_1, t_2]$ in $C_k$
$$\pi_k[\alpha^{t_{ki}}] \leftarrow \pi_k[\alpha^{t_{ki}}] \cdot \frac{\delta_{j \to k}[\alpha^{t_{ki}}]}{\mu_{j,k}^l[\alpha^{t_{ki}}]}$$
$$\pi_k[\beta^{t_{k(i+1)}}] \leftarrow \pi_k[\beta^{t_{k(i+1)}}] \cdot \frac{\delta_{j \to k}[\beta^{t_{k(i+1)}}]}{\mu_{j,k}^l[\beta^{t_{k(i+1)}}]}$$
$$\pi_k[\mathbf{Q}^{t_{ki}}] \leftarrow \pi_k[\mathbf{Q}^{t_{ki}}] + \lambda(\delta_{j \to k}[\mathbf{Q}] - \mu_{j,k}^l[\mathbf{Q}])$$
4. $\mu_{j,k}^l \leftarrow \delta_{j \to k}$

Note that we scale the update of **Q** by $\lambda$. This is so because sometimes the update may lead to a **Q** matrix that has negative off-diagonal values which is not admissable by our definition of a valid intensity matrix. This problem is not peculiar to Dynamic-EP. A similar problem is encountered in Gaussian-EP (Minka, 2001). Hence, we find the largest $\lambda$ such that the updated **Q** matrix is valid. This change does not affect the fixed-point of the algorithm. At convergence (i.e., when $\delta_{j \to k}[\mathbf{Q}]$ matches $\mu_{j,k}^l[\mathbf{Q}]$) this algorithm has the same fixed-point as the original algorithm.

**Procedure** *Send-Horizontal-Message*$(i, j, S_{ij}^1)$

1. $t \leftarrow \mathcal{I}_{ij}^1$
1. $[t_{i1}\ t_{i2}] \leftarrow \mathcal{I}_i$
2. $[t_{j1}\ t_{j2}] \leftarrow \mathcal{I}_j$
3. $\delta_{i \to j}[\alpha^t] \leftarrow \pi_i[\alpha^t],\ \delta_{i \to j}[\beta^t] \leftarrow \pi_i[\beta^t]$
4. If $(t_{i2} = t_{j1} = t)$
$$\pi_j[\alpha^t] \leftarrow \pi_j[\alpha^t] \cdot \frac{\delta_{i \to j}[\alpha^t]}{\mu_{i,j}^1[\alpha^t]}$$
Else If $(t_{i1} = t_{j2} = t)$
$$\pi_j[\beta^t] \leftarrow \pi_j[\beta^t] \cdot \frac{\delta_{i \to j}[\beta^t]}{\mu_{i,j}^1[\beta^t]}$$
5. $\mu_{i,j}^1 \leftarrow \delta_{i \to j}$

**Procedure** *Update-Dist*$(j)$

1. Foreach consecutive sub-interval
$(t_{j(i-1)}, t_{ji})$ and $(t_{ji}, t_{j(i+1)})$ in $C_i$
$$\delta[\alpha] \leftarrow \alpha^{t_{j(i-1)}} \exp(\mathbf{Q}^{t_{j(i-1)}}(t_{ji} - t_{j(i-1)})))$$
$$\pi_j[\alpha^{t_{ji}}] \leftarrow \pi_j[\alpha^{t_{ji}}] \cdot \frac{\delta[\alpha]}{\mu_i^{t_{ji}}[\alpha]}$$
$$\mu_i^{t_{ji}}[\alpha] \leftarrow \delta[\alpha]$$
2. Foreach consecutive sub-interval
$(t_{ji}, t_{j(i+1)})$ and $(t_{j(i-1)}, t_{ji})$ in $C_i$
$$\delta[\beta] \leftarrow \exp(\mathbf{Q}^{t_{ji}}(t_{j(i+1)} - t_{ji}))\beta^{t_{j(i+1)}}$$
$$\pi_j[\beta^{t_{ji}}] \leftarrow \pi_j[\beta^{t_{ji}}] \cdot \frac{\delta[\beta]}{\mu_i^{t_{ji}}[\beta]}$$
$$\mu_i^{t_{ji}}[\beta] \leftarrow \delta[\beta]$$

For updating the distribution within a cluster, we have the flexibility not to fully calibrate the nested cluster graph whenever we receive a message; for example, we can update a subset of the sub-interval factors as needed for the vertical message computations, saving computational cost. Alternatively, we can incorporate multiple incoming messages before recalibrating.

## 5 Dynamic Repartitioning of Messages

The message over the variables $V_{jk}$ of sepset $S_{j,k}$ is constrained to belong to the set of homogeneous Markov Processes, so that a single intensity matrix is used to describe the evolution of the variables over the interval $\mathcal{I}_{jk}$. Additional partitioning of $\mathcal{I}_{jk}$ allows us to have a richer piecewise homogeneous representation over the same interval.

To dynamically change the granularity of messages, we consider (online during inference) the possibility of splitting $S_{j,k}$ into sepsets $S_{j,k}^l$ over sub-intervals $\mathcal{I}_{jk}^l$, thereby creating new demarcation points. Two questions naturally arise: how should we choose where these new demarcation points should go and how should we decide whether or not to make any particular split? The answer to these questions is not obvious. We might base the decision of cluster splits on the order of magnitude of the diagonal elements in the cluster intensity matrices, as large values in the intensity matrix mean a faster rate of evolution. This may be a good heuristic, but it takes into account neither the starting distribution (which has a significant impact on the relevance on any intensity) nor evidence received as messages from neighbors.

We provide an alternative approach, which adds one split point at a time, and makes use of the KL-divergence minimization (i.e., projection) that we must perform anyway. The basis for our analysis is the following result, which follows from standard results for the exponential family:

**Proposition 5.1** *Let $P_C$ be the distribution over variables $V$ of cluster $C$ for the interval $\mathcal{I}_C$ defined by parameters $\eta_C$. Let $P_S$ be a distribution over variables $V' \subseteq V$ of sepset $S$ for the interval $\mathcal{I}_S \subseteq \mathcal{I}_C$ defined by parameters $\eta_S$. If $\mathbf{E}_C[\tau(V)]$ is the expected sufficient statistics over variables $V$ as computed from $P_C$, then*

$$\mathbf{D}(P_C || P_S) = \langle \mathbf{E}_C[\tau(V)], \eta_C \rangle - \langle \mathbf{E}_C[\tau(V')], \eta_S \rangle - \ln \frac{Z_C}{Z_S},$$

*where $\mathbf{E}_\mathcal{C}[\tau(\mathbf{V}')]$ is computed by marginalization.*

Consider splitting sepset $\mathcal{S}$ at time $t$. This would give us two homogeneous approximations $P_{\mathcal{S}_L}$ and $P_{\mathcal{S}_R}$ computed as described in the previous section. Using Propositions 5.1 we can compute the "cost" (in terms of KL divergence) of using a two-piece approximation rather than the correct cluster distribution, $C^2_{KL} = \mathbf{D}(P_\mathcal{C}||P_{\mathcal{S}_L}) + \mathbf{D}(P_\mathcal{C}||P_{\mathcal{S}_R})$. We can similarly define the cost for the single-piece approximation: $C^1_{KL} = \mathbf{D}(P_\mathcal{C}||P_\mathcal{S})$. We would like to select the repartition point $\hat{t} = \arg\min_t C^2_{KL}(t)$.

It turns out that we can perform this computation efficiently by building on properties of the fifth-order Runge-Kutta, which is used as the key subroutine in the inference process. The Runge-Kutta method uses an adaptive parameter to decide the step size based on the size of the errors accumulated while performing the integration. Hence, on intervals where the errors are large, it takes smaller steps and vice versa. At all the interval partitions formed by the Runge-Kutta points, we can compute the $\hat{\mathbf{Q}}_L$ and $\hat{\mathbf{Q}}_R$ by computing the intensity matrices incrementally from the sufficient statistics over the left and right sub-intervals. This is an $\mathcal{O}(n^2)$ computation where $n$ is the dimension of the intensity matrix and each step of Runge-Kutta is $\mathcal{O}(n^3)$. Given the intensity matrices, and removing terms that remain constant over different partitions, the $C_{KL}$ computation simplifies to only the terms involving inner products over marginalized expected sufficient statistics and the partition functions (when there is continuous evidence). Each of the inner-products are over vectors of length $d^2$, where $d$ is the dimension of the intensity matrix over the sepset variables and $d < n$. The Runge-Kutta computations are performed at $\mathcal{O}(\hat{q}T)$, where $\hat{q}$ is the maximum intensity value in $\mathbf{Q}$. Hence, this optimization is $\mathcal{O}(\hat{q}d^2T)$.

In order to decide whether to actually split at $\hat{t}$, we compare the KL cost of using the one-piece approximation to the two-piece. That is, we define a threshold, $k^*$ and make the split if $C^1_{KL} - C^2_{KL} > k^*$. Again, this computation can be incorporated efficiently given that we have already stored the incremental sufficient statistics. After the 2-piece message has been computed, to incorporate the message in the receiving cluster, a demarcation point is first created at $\hat{t}$. Then, each sepset message is incorporated independently.

## 6 Results

We use an extended version of the drug effect network of NKS shown in Fig. 1 to illustrate the dynamic behavior of this algorithm. The network models the effects of the uptake of a drug and the resulting concentration of the drug in the bloodstream.

**Illustrative Examples.** In our first scenario, we model a person that is experiencing joint pain, takes the drug to alleviate the pain, is not eating, has an empty stomach, is not hungry, and is not drowsy. The weather condition is nor-mal and the barometric pressure is steady. In Fig. 2(a), we show part of the cluster graph with the sepsets created as a result of message passing. The clusters have been unrolled over a period of 10 hours. $\mathcal{C}_0$ which contains variables such as *Hungry*, *Eating* and *Full-Stomach* changes very rapidly in the first hour because the empty stomach leads to hunger which causes the person to eat. This behavior results in **Dynamic-EP** creating several splits to refine the granularity at which the message is sent to $\mathcal{C}_1$ within the first few hours. $\mathcal{C}_4$, on the contrary, evolves very slowly when the weather is steady, a state with relatively high persistence. As a result, this causes no bad effects on the barometric pressure and it continues to remain steady. As expected, $\mathcal{C}_4$ and $\mathcal{C}_2$ do not further refine the message they exchange over the variable *Barometer Pressure*. To examine the effect of incoming evidence, in our second scenario, we now observe very bad weather. This adversely makes the barometric pressure unsteady. $\mathcal{C}_4$ now dynamically splits its message to 8 separate pieces as shown in Fig. 2(b).

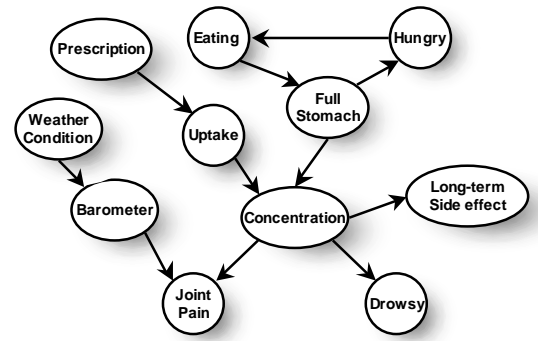

Figure 1: The drug-effect network

**Dynamic-EP** also shows interesting behavior with regard to deciding when to partition a sepset. In this split graph, $\mathcal{C}_0$ first creates a partition at time 2.1108 to the message it sends to $\mathcal{C}_1$ over $\mu_{0,1}$. $\mathcal{C}_1$ emulates the behavior by creating a partition to the message it sends to $\mathcal{C}_2$ at time 2.1108. $\mathcal{C}_1$ in future iterations of message passing chooses not to further refine its message over $\mu_{1,2}$ over time $0 - 2.1108$ until it receives a message from $\mathcal{C}_0$ split at time $0.489$ after which it splits its message over $\mu_{1,2}$ from time $0 - 2.1108$ at $0.472$. Thus, the splitting behavior happens selectively depending on how the incoming message affects the message computation at a cluster given its own evolution matrix. The EP algorithm of NKS is unable to respond similarly to an incoming message – once a granularity of computation at a given time has been selected for the entire system, it cannot adaptively refine its messages as needed.

**Quantitative Results.** To investigate the computational properties of our dynamic-EP algorithm, we compared its performance to the EP algorithm of NKS, using various uniform time granularities. Unfortunately, the extended version of the drug effect network is too large to allow ex-act inference for comparison. We therefore generated chain CTBNs $X_1 \to \ldots \to X_n$ of different lengths $n$, ranging from 5 to 50 variables, in increments of 5. Each variable $X_i$ has three values. A child $X_{i+1}$ generally tries to follow the value of its parent $X_i$, with some noise; $X_{i+1}$ transitions an order of magnitude faster than $X_1$ when it disagrees with $X_i$, and an order of magnitude slower than $X_1$ when it agrees with $X_i$. We use trajectories of 10 time units in length, and compared four different approximation schemes: **Uniform-K** — splitting clusters uniformly, as in the NKS algorithm, with time granularity $K = 1, 5, 10$; and our dynamic-EP algorithm with a KL threshold of 0.01. Note that, at uniform granularity 10, there are no splits. In all cases, our cluster graph contained uniform clusters, consisting of pairs $\{X_i, X_{i+1}\}$.

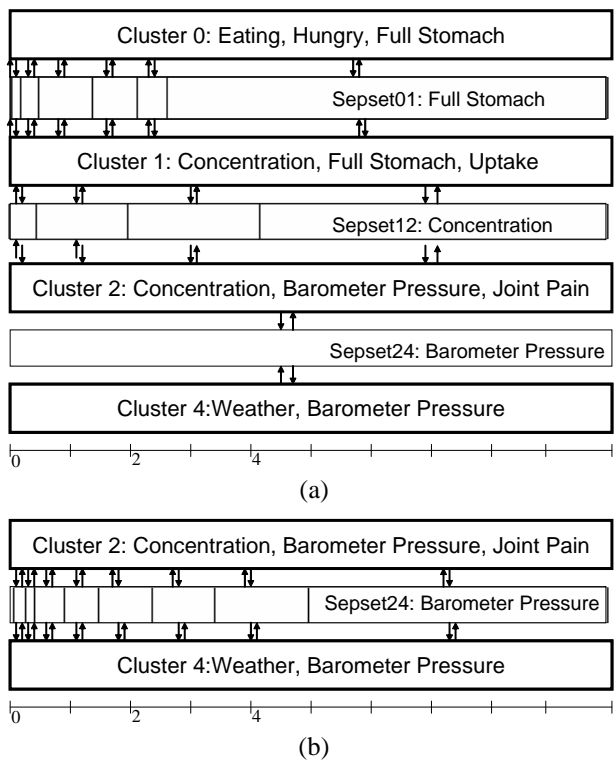

Figure 2: (a) Portion of drug-effect network cluster graph pulled out to show dynamically selected partition points under the joint pain scenario.. (b) Portion of cluster graph showing partition points for the bad weather scenario.

We considered a scenario where we have evidence only at the initial time. Thus the initial state is unstable and results in rapid change. To capture this phenomenon, we need a finer-grained approximation for a short duration, then a coarser approximation as the distribution approaches equilibrium. Fig. 3(a) presents results for a distribution over the 5-variable chain, which is small enough to admit a comparison to exact inference. The graph plots the KL-divergence between the exact distribution and that given by four approximate methods for 100 points in $[0, 10]$. Dynamic-EP generated only a single split point, in the sepset between the cluster $\{X_1, X_2\}$ and $\{X_2, X_3\}$, at time point 0.762. With the addition of this one sepset, Dynamic-EP does as well or better than Uniform-1, except in a very small interval, where it does almost as well. In terms of running time, Uniform-1 took 0.67s, Uniform-5 took 0.14s, Uniform-10 took 0.05s, and Dynamic-EP took 0.08s, only slightly more than Uniform-10. The threshold $k^*$ in Dynamic-EP controls the accuracy versus complexity of computation tradeoff. Lowering this threshold would create more splits and improve the accuracy in the small interval where it does not do as well at increased computational cost.

To obtain results for larger networks where exact inference is intractable, we used an empirical approximation to the KL-divergence. For each network, we generated 100 random trajectories from the network, and ran approximate inference for each one. Then, for each $X_i$ and each of 100 time points $t \in (0, 10]$, we computed the log-likelihood for the true value $x_i^t$ of $X_i^t$ in the trajectory, using the marginal distribution of $X_i^t$ in a cluster that contains it. We averaged the log-likelihood over variables and trajectories. The results, graphed in Fig. 3(b), show that the performance of Dynamic-EP is only slightly worse than those of Uniform-1. Again, this outcome can be changed by using a lower KL-threshold. Moreover, in contrast to the coarser uniform partitioning algorithms, it degrades much more slowly as the number of variables increases. Fig. 3(c) shows the average running time taken by each of the approximate methods, showing that this high accuracy is obtained at a computational cost which is comparable to that of Uniform-10.

Since Dynamic-EP can focus computational resources on portions of the cluster graph that are evolving faster, we wanted to explore the speed-up we achieve in networks where there is widening gap between the rate of the fastest evolving cluster and the slowest. So, using a 30 variable chain whose top node evolved at a fast rate (max intensity = 100), we made a series of networks by slowing the evolution rate of the remaining variables — leading to ratios from 1 (i.e., all clusters evolve at the same rate) to $10^4$ (i.e., the cluster containing the top node evolves $10^4$ times faster than the others). Fig. 3(d) shows the resulting speed-up expressed as the ratio of Uniform-0.1 runtime over Dynamic-EP plotted against the series of networks with increasingly divergent rates of evolution between the fastest and slowest clusters. This graph represents the average speed-up over 10 runs. As we expect, the figure shows an increasing advantage for Dynamic-EP over Uniform-0.1 when the clusters evolve at increasingly different rates. There is a peak at cluster rate ratio is 200 but the errorbar shows a high range of values at that point possibly due to interactions with the particular KL threshhold of the runs (0.01).

## 7 Discussion and Future Work

We have presented a highly flexible cluster graph architecture for passing messages across both time and space in CTBNs. We also presented **Dynamic-EP**, a new algorithm for approximate inference in CTBNs. This algorithm adaptively assigns computational resources to parts of the inference where greater accuracy is required, and can provide a much better tradeoff between computational cost and accuracy than previous algorithms. Most importantly, Dynamic-EP deals well with situations where some components of the system evolve much more rapidly than others, allowing each part of the system to adaptively choose the time granularity most appropriate to it at that time.

There are many useful extensions of this work. Clearly, we plan to test whether the computational gains on simple, synthetic networks also manifest in real-world problems. More broadly, our framework allows a highly flexible inference architecture, where process variables can dynamically change their cluster assignments over time. Thus, if two variables undergo a strong interaction, we can temporarily put them in the same cluster. It would be interesting to design an algorithm that dynamically determined an appropriate cluster structure as the process evolves. Finally, there are many probabilistic models other than CTBNs where EP is used to provide a parametric approximation to complex messages in a cluster graph. In some cases, there may be a need for a richer, more flexible representation of the messages (one of the key motivations for the development of non-parametric belief propagation (Sudderth et al., 2003).) The algorithm that we proposed provides a semi-parametric message representation. It would be interesting to explore the viability of a similar approach in other types of probabilistic graphical models.

## References

Dean, T., & Kanazawa, K. (1989). A model for reasoning about persistence and causation. *Computational Intelligence*, *5*, 142–150.

Kjaerulff, U. (1997). Nested junction trees. *UAI*.

Lauritzen, S. (1996). *Graphical models*. Clarendon Press.

Minka, T. (2001). Expectation propagation for approximate bayesian inference. *UAI* (pp. 362–369).

Nodelman, U., Koller, D., & Shelton, C. (2005). Expectation propagation for continuous time Bayesian networks. *UAI*.

Nodelman, U., Shelton, C., & Koller, D. (2002). Continuous time Bayesian networks. *UAI* (pp. 378–387).

Norris, J. (1997). *Markov chains*. Cambridge Univ. Press.

Rabiner, L. R., & Juang, B. H. (1986). An introduction to hidden Markov models. *IEEE ASSP Magazine*, 4–16.

Sudderth, E., Ihler, A., Freeman, W., & Willsky, A. (2003). Nonparametric belief propagation. *CVPR*.

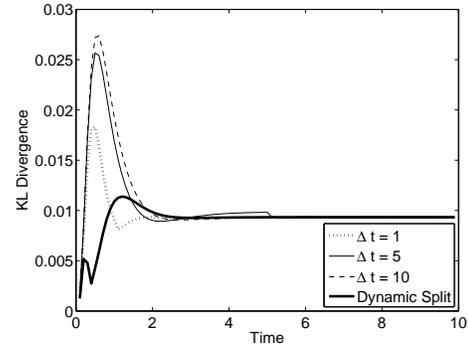

(a)

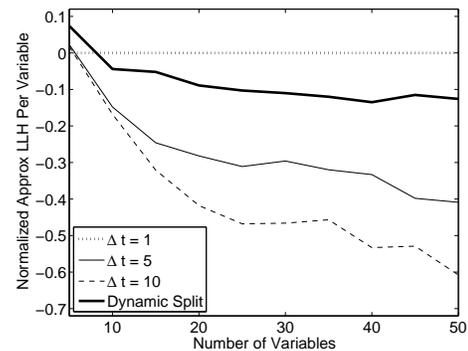

(b)

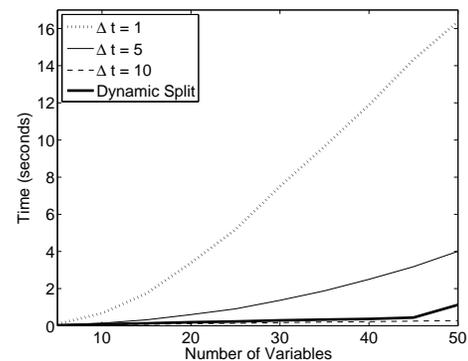

(c)

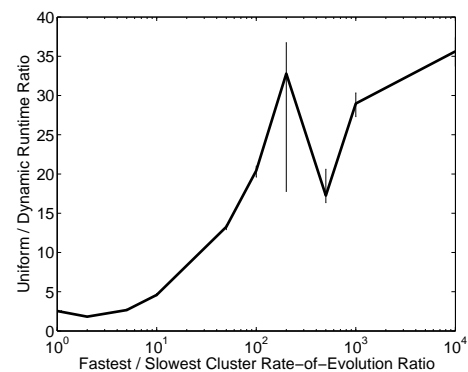

(d)

Figure 3: (a) KL from exact distribution for 5-chain. (b) Approx LLH per variable in chains of increasing length. (c) Processor time to run approximate inference for chains of increasing length. (d) Speed-up of Dynamic-EP over Uniform-0.1 for increasing difference between rate of fastest and slowest clusters